\pdfoutput=1
\documentclass[11pt]{article}

\usepackage[]{EMNLP2022}

\usepackage{times}
\usepackage{latexsym}
\usepackage{adjustbox}
\usepackage{enumitem}
\usepackage{multirow}
\usepackage[T1]{fontenc}
\usepackage[utf8]{inputenc}

\usepackage{microtype}

\usepackage{inconsolata}

\usepackage{xcolor}

\DeclareMathAlphabet\mathbfcal{OMS}{cmsy}{b}{n}
\usepackage{mathtools}
\usepackage{algorithm}
\usepackage{algpseudocode}
\usepackage{physics}
\usepackage{subeqnarray}
\usepackage{amsfonts}
\usepackage{amsmath,amssymb}
\DeclarePairedDelimiterX{\infdivx}[2]{(}{)}{%
  #1\;\delimsize\|\;#2%
}

\usepackage{capt-of}

\title{Hardness-guided domain adaptation to recognise biomedical named entities under low-resource scenarios}

\author{
    Ngoc Dang Nguyen \\
    Monash University\\
    \texttt{dan.nguyen2@monash.edu} \\\And
    Lan Du\thanks{~~Corresponding Author}\\
    Monash University\\
    \texttt{lan.du@monash.edu} \\\AND
    Wray Buntine \\
    VinUniversity\\
    \texttt{wray.b@vinuni.edu.vn} \\\And
    Changyou Chen \\
    University at Buffalo \\
    \texttt{changyou@buffalo.edu} \\\And
    Richard Beare \\
    Monash University\\
    \texttt{richard.beare@monash.edu} \\    
    }

\begin{document}
\maketitle
\begin{abstract}%transferring knowledge from high resources to low-resources biomedical domain.
Domain adaptation is an effective solution to data scarcity in low-resource scenarios. However, when applied to token-level tasks such as bioNER, domain adaptation methods often suffer from the challenging linguistic characteristics that clinical narratives possess, which leads to unsatsifactory performance.
% Low resource biomedical learning has been a big challenge for many classification tasks, where the final classifier is trained only with a few examples. 
% This problem amplifies when we apply the few-shot setup to recognising named entity from different domains, i.e., few-shot domain adaption for NER.
In this paper, we present a simple yet effective hardness-guided domain adaptation (HGDA) framework for bioNER tasks that can effectively leverage the domain hardness information to improve the adaptability of the learnt model in the low-resource scenarios.
Experimental results on biomedical datasets show that our model can achieve significant performance improvement over the recently published state-of-the-art (SOTA) MetaNER model.
\end{abstract}
%%%%%%%%%%%%%%%%%%%%%%%%%%%%%%%%%%%%%%%%%%%%
\section{Introduction}
Named Entity Recognition (NER) is a fundamental NLP task which aims to locate named entity (NE) mentions and classify them into predefined categories such as location, organization, or person. 
NER usually serves as an important first sub-task for
%information extraction \cite{ritter2012open},
information retrieval \cite{banerjee2019information}, task oriented dialogues \cite{peng2020soloist} and other language applications. 
Consequently, NER has seen significant performance improvements with the recent advances of pre-trained language models (PLMs) \cite{akbik-etal-2019-pooled,devlin-etal-2019-bert}. 
Unfortunately, a large amount of training data is often essential for these PLMs to excel and except for a few high-resource domains, the majority of domains have limited amount of labeled data.

This data-scarcity problem amplifies given the context of biomedical NER (bioNER).
Firstly, the annotation process for the biomedical domains is time-consuming and
%requires medical expert experience which 
can be extremely expensive.
Thus, many biomedical domain corpora, especially for those privately developed, are often scarcely labeled. 
%This prevents the application of the mentioned PLMs. 
Furthermore, each biomedical domain can have distinct linguistic characteristics which are non-overlapping with those found in other biomedical domains \cite{10.1093/bioinformatics/btz682}. 
This linguistic challenge often diminishes the robustness of PLMs transferred from high-resource biomedical domains to low-resource ones \cite{10.1093/bioinformatics/btz504}. 

Given the premise, this paper focuses on adapting PLMs for bioNER tasks learned from high-resource biomedical domains to low-resource ones.
A potential solution is to inject the prior ``experience'' to this adaptation process.
Few works have explored this area such as \citet{li2020few} and \citet{10.1145/3366423.3380127}, 
where the former followed the optimization/meta-learning strategy by \cite{finn2017model} 
and the latter introduced a feature critic module similar to the work of \citet{pmlr-v97-li19l}.

We show that simply incorporating the hardness information that each domain contributes to the learning of the bioNER LMs could significantly boost the adaptation performance of existing learning paradigms under various low-resource settings. 
This happens since the importance/hardness of biomedical domains can vary significantly, as shown in \autoref{tab:tasks}.
While some domains might contain a lot of NEs, many do not, ({\it e.g.}, the last row in \autoref{tab:tasks}), and hence, contribute little to learning the bioNER LMs.
Meanwhile, the domain difficulty ties both to the number of entities and to the length of those entities. Given the non-overlapping linguistic characteristics found in biomedical domains, this poses another challenge to effectively adapt the trained bioNER LMs to the new domain.
Therefore, we argue that the current adaptation framework proposed by \citet{10.1145/3366423.3380127} could be further enhanced using the hardness information.
%according to their hardness in order to improve learning efficiency and the model performance.
We present two simple but effective ways of incorporating hardness information into our learning framework, named HGDA.
We show that our hardness-guided domain adaptation approaches for bioNER tasks outperform the SOTA domain adaptation NER technique by \citet{10.1145/3366423.3380127}.
% Including transformers

\begin{table*}[t]%&{0.}&{0.}&{0.}&{0.}&{0.}&{0.}&{0.}&{0.}&{0.}\\
\centering
\begin{adjustbox}{max width=\textwidth}
\begin{tabular}{|l|c|}
    \hline
    \multicolumn{1}{|c|}{\textbf{Example}} 
    &\multicolumn{1}{c|}{\textbf{Score}} 
    \\ \hline 
    
    Stimulation of human neutrophils with \textbf{\color{red} [chemoattractants]} \textbf{\color{red}[FMLP]} or \textbf{\color{red}[platelet activating factor (PAF)]} results in different but overlapping functional responses.
    &{0.46}\\
    
    Of even more interest, \textbf{\color{red}[IkappaBalpha]} overexpression inhibited the production of \textbf{\color{red}[matrix metalloproteinases 1 and 3]} while not affecting their tissue inhibitor.    
    &\\
    
    ...more durable inhibition of HIV - 1 replication than was seen with the \textbf{\color{red}[NF-kappa B]} inhibitors alone or the \textbf{\color{red}[anti-Tat sFv intrabodies]} alone.
    &\\

    Spontaneous occurrence of early region 1A reiteration mutants of type 5 adenovirus in persistently infected human T-lymphocytes.
    &\\
    \hline
    
    Here we report the fabrication of single-molecule transistors based on individual C60 molecules connected to gold electrodes.
    &{0.18}\\
    
    The contractile effects of \textbf{\color{red}[oxytocin]}, prostaglandin F2 alpha and their combined use on human pregnant myometrium were studied in vitro.
    &\\
    
    Transcriptional activation of the \textbf{\color{red}[proopiomelanocortin gene]} by \textbf{\color{red}[cyclic AMP-responsive element binding protein]}.
    &\\
    
    The difference between the effects of the two dose levels of Z.
    &\\
    \hline
    
    She was monitored for one more day and then discharged with instructions to discontinue her diet pills    
    &{0.01}\\

    The Raf/Ras/ERK/MAPK pathway is known to be involved in NGF-induced outgrowth
    &\\
    
    Our analysis reveals that the oviduct is lined, along its entire length, by a monolayered epithelium comprised of squamous-type cells.
    &\\
    
    In one case study, Bramson et al.
    &\\
    
    \hline
    \end{tabular}
\end{adjustbox}
% \vspace{-3mm}
\caption{
Examples of domain hardness scores (computed from our method) for tasks generated from three domains (gene, drug, and species respectively) during the training procedure. 
The score is based on a scale from 0 to 1, the higher the score, the more challenging the domain is. The NEs are put in brackets with red color for each sentence.
}
\label{tab:tasks}
% \vspace{-6mm}
\end{table*}
%%%%%%%%%%%%%%%%%%%%%%%%%%%%%%%%%%%%%%%%%%%%
\section{Related Works}
% \vspace{-2mm}
Few works have addressed domain adaptation for NER. 
Both \citet{li2020few} and \citet{10.1145/3366423.3380127} seek a robust representation for the sequence labeling function BiLSTM-CRF using the meta-learning framework \cite{finn2017model},
with the latter further includes an auxiliary network to promote adversarial learning.
%during the training process.
\citet{hu-etal-2022-label} consider label dependencies via an auto-regressive framework built on top of Bi-LSTM for cross-domain NER.
However, different domains can have different level of hardness which both works have not yet addressed.
Existing domain adaptation techniques using meta learning framework to incorporate hardness information via
1) actively ranking the tasks in term of difficulty level~\cite{yao2021meta,zhou2020expert,liu2020adaptive,achille2019task2vec};
2) designing an adaptive task scheduler \cite{yao2021meta};
or 
3) relying on generative approaches to quantify the uncertainties of tasks~\cite{kaddour2020probabilistic,nguyen2021probabilistic}.
To our knowledge, we are the first to perform hardness guided domain adaptation for bioNER tasks.

% \vspace{-3mm}
%%%%%%%%%%%%%%%%%%%%%%%%%%%%%%%%%%%%%%%%%%%%
\section{Hardness-guided domain adaptation}
% \vspace{-2mm}
\subsection{Problem Setup} 
Given a set of biomedical corpora from multiple source domains $\mathcal{D}_{\text{source}}$ ({\it e.g.}, Drug, Gene, Species, etc), we aim to learn a sequence labelling function $h: \mathcal{X}\rightarrow\mathcal{Y}$
\footnote{
$\mathcal{X}=\{x^i_1,\ldots,x^i_L\}_{i=1}^{N}$, and
$\mathcal{Y}=\{y^i_1,\ldots,y^i_L\}_{i=1}^{N}$.
$\mathcal{X}$ and $\mathcal{Y}$ denote the set of sentences and tags/labels respectively. $N$ is the total number of sentences, 
and $L$ is the number of word tokens for the sentence $i$.
} 
from a set of tasks $p(\mathcal{T})$ sampled from $\mathcal{D}_{\text{source}}$ so that $h$ can be adapted to a new task $\mathcal{T}^{'}$ sampled from the target domain $\mathcal{D}_{\text{target}}$ (e.g., Disease). 
This function $h$ should contain 
1) a sentence encoder parameterized with  $\vb*\theta$ ({\it e.g.}, BiLSTM) that captures the contextual information about words, 
and 
2) a tag decoder parameterized with $\vb*\phi$ ({\it e.g.}, CRF) that assigns the entity tags to these words
\footnote{
We consider the BIO tagging schema. %containing three labels: B-Begin, I-Inside, and O-Outside.
}.
Thus, the learning objective is to search for the optimal $\vb*\Theta^*\equiv\{\vb*\theta, \vb*\phi\}$ from $\mathcal{D}_{\text{source}}$.
This optimal $\vb*\Theta^*$ should minimise the risk of adapting $h$ from $\mathcal{D}_{\text{source}}$ to $\mathcal{T}^{'}$ from $\mathcal{D}_{\text{target}}$.

\subsection{Task Generation}
To optimize for $\vb*\Theta^*$ with stochastic optimization, one first needs to sample from $p(\mathcal{T})$, {\it i.e.}, task generation. 
Each bioNER task $\mathcal{T}_i$ in our setting is divided into a support set $\mathcal{T}_i^S$ and a query set $\mathcal{T}_i^Q$, with $\mathcal{T}_i^S\cap \mathcal{T}_i^Q=\emptyset$. 
We further restrict both $\mathcal{T}_i^S$ and $\mathcal{T}_i^Q$ to contain only $K$ sentences respectively sampled from a domain in $\mathcal{D}_{\text{source}}$. This value of $K$ is dependent on the amount of data we have during adaptation phase for $\mathcal{T}^{'}$ and can be as small as 5 or 10.
This is to mimic the same few-shot setting in the training phase which has been shown to reduce the PAC-Bayesian error bound during the adaptation phase \cite{ding2021bridging}.
To encode the hardness information into our task generation process, we further consider the imbalance issue caused by the NER tasks.
As shown in \autoref{tab:Corpora}, the majority of the sentences in the biomedical corpora does not contain any NEs.
Thus, it is highly likely that the $K$ randomly-sampled sentences contain no NEs, which can result in a biased sequence labeller that always predicts ``O'' in the adaption phase.
To avoid this issue, we propose our first HGDA approach by selecting the $K$ sentences in $\mathcal{T}_i^S$ to be those containing at least one biomedical NE, which is shown to be highly effective during the adaptation phase.

\subsection{Bilevel Optimization}
To regularize $\vb*\theta$, HGDA includes a domain classifier as a separate head on top of the sentence encoder.
This enforces the network to learn a domain conditional invariant sentence encoder \cite{blanchard2017domain,li2018deep,shao2019multi}.
This domain classifier, parameterized with $\vb*\omega$, consists of a fully connected layer and is used to predict which domain the sentences in a task $\mathcal{T}_i$ belong to.
The classification function $f$ will henceforth be used to represent the composition of the sentence encoder and the domain classifier.
Consequently, the learning objective of HGDA is
\begin{equation}
    \small
    %\label{eq:MetaLoss}
    \mathcal{L}_i=
    \mathcal{L}^{\text{lab}}\left(h\left(\vb*\theta, \vb*\phi\right),\mathcal{T}_i\right)+
    \lambda\mathcal{L}^{\text{cls}}\left(f\left(\vb*\theta, \vb*\omega\right),\mathcal{T}_i\right),
\end{equation}
where $\lambda$ control the trade-off between the labelling loss and the classification loss.
As HGDA follows the bilevel optimization framework, we first generate a batch of task from $p(\mathcal{T})$.
For each $\mathcal{T}_i$ in this batch, we train the model on $\mathcal{T}_i^S$ then validate the performance on $\mathcal{T}_i^Q$ using our learning objective.
Consequently, we gather the gradients from each $\mathcal{T}_i$ in the current batch of task and make the update to the parameters, finishing one iteration of the training process.
This runs until no further improvement can be made. The full algorithm is summarized by Alg.~\ref{alg:MetaBioNER} and Alg.~\ref{alg:Loss} in the appendix.

\subsection{Task Hardness}
Although picking the $K$ sentences with NEs for $\mathcal{T}_i$ is shown to improve the DA performance (see \autoref{tab:heterogeneous}), 
it is not realistic in practice to have only sentences with NEs and wasteful not using the sentences without NEs as these sentences would still provide the sentence encoder with important contextual information of the clinical narratives.
Hence, HGDA incorporates another simple but effective way of computing the bioNER task hardness based on the losses.
The gradients propagated by $\mathcal{T}_i$ will be weighted by the hardness level of $\mathcal{T}_i$.
Specifically, we define the task difficulty $\Gamma_i = \{\gamma_i^{\theta}, \gamma_i^{\phi}, \gamma_i^{\omega}\}$ for task $\mathcal{T}_i$ with its corresponding objective values as follows
% {\small
% \vspace{-3mm}
\begin{equation}
    \small
    \label{eq:TaskScore}
    % \begin{align}
        \gamma_i^{\theta} = \frac{\mathcal{L}_i}{\sum\mathcal{L}_j};\;
        \gamma_i^{\phi} = \frac{\mathcal{L}_i^{\text{lab}}}{\sum\mathcal{L}^{\text{lab}}_j};\;
        \gamma_i^{\omega} = \frac{\mathcal{L}_i^{\text{cls}}}{\sum\mathcal{L}^{\text{cls}}_j}~,
    % \end{align}
% \vspace{-3mm}
\end{equation}
% }
where $\{\gamma_i^{\theta}, \gamma_i^{\phi}, \gamma_i^{\omega}\}$ represent the task hardness scores to update $\{\theta, \phi, \omega\}$ respectively.
By incorporating task hardness in the optimization process, 
HGDA, after collecting adequate contextual information for the sentence encoder, should gradually shift the focus to more challenging labelling tasks for the tag-decoder rather than the ones that contribute little to no learning value, {\it e.g.}, a task that contains short and simple sentences without bioNEs. 
This happens as multiplying the hardness score with the corresponding gradient value will force the gradient update to zero for sentences with no NEs.
\autoref{tab:tasks} shows how HGDA ranks the contribution of each task towards the gradient updates.

\begin{table}[!t]
    \tiny
    \centering
    \begin{tabular}{|l|l|c|c|}
            \hline
            \multicolumn{1}{|c|}{\textbf{Corpora}} &
            \textbf{\begin{tabular}[l]{@{}c@{}}Entity\\ Type\end{tabular}} &
            \textbf{\begin{tabular}[c]{@{}c@{}}No. Unique\\ Tokens\end{tabular}} &
            \textbf{\begin{tabular}[c]{@{}c@{}}\% sentences\\ with NEs\end{tabular}} 
            \\ \hline
            NCBI~\cite{10.5555/2598938.2599127}         & Disease      & $12,128$   &55    \\ \hline
            BC5CDR~\cite{article8}                      & Disease      & $23,068$   &59    \\ \hline
            BC5CDR~\cite{article8}                      & Drug    & $23,068$   &65    \\ \hline
            BC4CHEMD~\cite{article9}                    & Drug    & $114,837$  &48    \\ \hline
            JNLPBA~\cite{collier-kim-2004-introduction} & Gene      & $25,046$   &81    \\ \hline
            BC2GM~\cite{article11}                      & Gene & $50,864$   &51    \\ \hline
            LINNAEUS~\cite{article10}                   & Species      & $34,396$   &13    \\ \hline
            S800~\cite{10.1371/journal.pone.0065390}    & Species      & $205,26$   &30    \\ \hline
        \end{tabular}%
        % \vspace{-2mm}
    \caption{Biomedical corpora used in our experiments \cite{10.1093/bioinformatics/btx228,10.1093/bioinformatics/btz682,DBLP:journals/corr/abs-1810-10566}.}
    \label{tab:Corpora}
    % \vspace{-5mm}
\end{table}
%%%%%%%%%%%%%%%%%%%%%%%%%%%%%%%%%%%%%%%%%%%%
\section{Experimental Results}
% \vspace{-3mm}
\subsection{Datasets}
We use the pre-processed version of the benchmark corpora (see Tab.~\ref{tab:Corpora}) which were used by the SOTA bioNER BioBERT \cite{10.1093/bioinformatics/btz682} and are publicly available at BioBERT's github website~\footnote{\url{https://github.com/dmis-lab/biobert}}.
These corpora are categorized into four non-overlapping biomedical domains, namely Disease, Drug, Gene and Species, each of which will serve as the target domain in our DA experiments.
When the sentence encoder is BiLSTM, HGDA uses BioWordVec embeddings pre-trained based on both PubMed database and clinical notes from MIMIC-III \cite{DBLP:journals/corr/abs-1810-09302, article21}.

\begin{table*}[!th]
\centering
\small
\begin{adjustbox}{max width=\textwidth}
\begin{tabular}{|c|l|cc|cc|cc|cc|c|}
    \hline
    \multicolumn{1}{|c|}{$\mathcal{T}^{'}$}
        & \multicolumn{1}{c|}{}
        & \multicolumn{2}{c|}{Disease} 
        & \multicolumn{2}{c|}{Drug} 
        & \multicolumn{2}{c|}{Gene} 
        & \multicolumn{2}{c|}{Species}
        & \multicolumn{1}{c|}{Overall}
    \\ \cline{3-10}
    
    \multicolumn{1}{|c|}{Size}
        & \multicolumn{1}{c|}{}
        & NCBI & BC5CDR 
        & BC5CDR & BC4CHEMD 
        & JNLPBA  & BC2GM
        & LINNAEUS  & S800 
        &    
    \\ \hline 
    
    \parbox[t]{3mm}{\multirow{6}{*}{\rotatebox[origin=c]{90}{\textbf{5}}}}
    &\textbf{MetaNER}
        &{0.2729}&{0.2171}&{0.5784}&{0.2212}&{0.2175}&{0.2443}&{0.1214}&{0.1516}&{0.2530}\\

    &\textbf{BioBERT}
        &{0.0428}&{0.0352}&{0.0600}&{0.0237}&{0.0727}&{0.0304}&{0.0081}&{0.0083}&{0.0352}\\
        
    &\textbf{HGDA}
        &{0.3001}&\textbf{0.2698}&\textbf{0.6102}&{0.2464}&{0.3687}&{0.3326}&\textbf{0.1753}&{0.2840}&\textbf{0.3234}\\
    
    &\textbf{HGDA-NEs}
        &{0.2825}&{0.2530}&{0.5517}&\textbf{0.2571}&\textbf{0.3776}&\textbf{0.3573}&{0.1557}&{0.2615}&{0.3121}\\

    &\textbf{HGDA*}
        &{0.2285}&{0.0678}&{0.4794}&{0.1288}&{0.3691}&{0.3226}&{0.0710}&{0.2563}&{0.2404}\\
    
    &\textbf{HGDA-NEs*}
        &\textbf{0.3125}&{0.1290}&{0.6066}&{0.2359}&{0.3298}&{0.3236}&{0.0701}&\textbf{0.2880}&{0.2869}\\
    
    \hline \hline
    
    \parbox[t]{3mm}{\multirow{6}{*}{\rotatebox[origin=c]{90}{\textbf{10}}}}
    &\textbf{MetaNER}
        &0.3330&0.3688&0.6659&0.3360&0.3374&{0.3265}&{0.3038}&{0.3164}&{0.3735}\\
    
    &\textbf{BioBERT}
        &{0.0905}&{0.0223}&{0.2315}&{0.0607}&{0.1961}&{0.2016}&{0.0162}&{0.0268}&{0.1057}\\
        
    &\textbf{HGDA}
        &{0.3953}&{0.4178}&{0.6798}&\textbf{0.4227}&\textbf{0.4790}&\textbf{0.4489}&\textbf{0.3201}%implicit {0.2939}
        &\textbf{0.3703}&\textbf{0.4417}\\

    &\textbf{HGDA-NEs}
        &\textbf{0.4386}&\textbf{0.4222}&{0.6605}&{0.3933}&{0.4371}&{0.4086}&{0.2474}&{0.3225}&{0.4163}\\

    &\textbf{HGDA*}
        &{0.3825}&{0.4014}&{0.6640}&{0.3566}&{0.4255}&{0.3974}&{0.1445}&{0.3631}&{0.3919}\\
    
    &\textbf{HGDA-NEs*}
        &{0.4084}&{0.3110}&\textbf{0.7097}&{0.4076}&{0.3966}&{0.3713}&{0.1228}&{0.3532}&{0.3851}\\
    
    \hline \hline
        
    \parbox[t]{3mm}{\multirow{6}{*}{\rotatebox[origin=c]{90}{\textbf{20}}}}
    &\textbf{MetaNER}
        &{0.4612}&{0.4722}&{0.7301}&{0.4383}&{0.4167}&{0.3926}&\textbf{0.4952}&{0.2977}&{0.4630}\\
    
    &\textbf{BioBERT}
        &{0.3296}&{0.2654}&{0.6225}&{0.2345}&{0.3751}&{0.4242}&{0.1004}&{0.2348}&{0.3233}\\
    
    &\textbf{HGDA}
        &\textbf{0.5631}&\textbf{0.5529}&\textbf{0.7472}&{0.4935}&\textbf{0.5466}&\textbf{0.5114}&{0.3657}&{0.4432}&{0.5280}\\

    &\textbf{HGDA-NEs}
        &{0.5540}&{0.5098}&{0.7305}&{0.4694}&{0.5375}&{0.5097}&{0.4843}&\textbf{0.5205}&\textbf{0.5394}\\
    
    &\textbf{HGDA*}
        &{0.4326}&{0.4703}&{0.7007}&{0.4494}&{0.4865}&{0.4356}&{0.1638}&{0.3694}&{0.4385}\\
    
    &\textbf{HGDA-NEs*}
        &{0.4789}&{0.5166}&{0.7340}&\textbf{0.4944}&{0.4694}&{0.4359}&{0.2859}&{0.4045}&{0.4775}\\
        
    \hline \hline
        
    \parbox[t]{3mm}{\multirow{6}{*}{\rotatebox[origin=c]{90}{\textbf{50}}}}
        
    &\textbf{MetaNER}
        &{0.5731}&{0.6106}&{0.7478}&{0.5082}&{0.5337}&{0.5058}&{0.6125}&{0.3607}&{0.5565}\\
    
    &\textbf{BioBERT}
        &{0.5998}&{0.5740}&{0.7520}&{0.4883}&{0.4855}&{0.5882}&{0.5835}&{0.4586}&{0.5662}\\  
    
    &\textbf{HGDA}
        &{0.6250}&{0.5939}&{0.7737}&{0.5728}&{0.5666}&{0.5442}&{0.6369}&\textbf{0.5855}&{0.6123}\\
    
    &\textbf{HGDA-NEs}
        &\textbf{0.6208}&{0.5847}&{0.7612}&{0.5781}&\textbf{0.6146}&\textbf{0.6016}&\textbf{0.6373}&{0.5445}&\textbf{0.6179}\\
    
    &\textbf{HGDA*}
        &{0.5618}&{0.5873}&{0.7584}&{0.5078}%...
        &{0.5256}&{0.4790}&{0.4526}
        &{0.4674}%...
        &{0.5425}\\
    
    &\textbf{HGDA-NEs*}
        &{0.6000}&\textbf{0.6190}&\textbf{0.8023}&\textbf{0.6273}%..
        &{0.5842}&{0.5464}&{0.4374}&{0.4678}&{0.5856}\\
    \hline
    \end{tabular}
\end{adjustbox}
% \vspace{-2mm}
\caption{
Average F1-performance of the sequence encoder adaptation for bioNER tasks with the best performance boldfaced. All results are averaged from 20 distinct samples, {\it e.g.}, given $\mathcal{T}^{'}$ size is 5, we adapt our HGDA variants and their baselines using $\mathcal{T}^{'}$ and validate their bioNER performance using the test data to record the f1-score. We then repeat this process with 20 different $\mathcal{T}^{'}$ of size 5, average the final results, and report the results using this table. \textbf{HGDA} and \textbf{HGDA-NEs} use BiLSTM as the sentence encoder, while \textbf{HGDA*} and \textbf{HGDA-NEs*} use BERT as the sentence encoder.
Unless otherwise specified, the HGDA-variants outperform their baselines with a p-value < 0.05.
}
\label{tab:heterogeneous}
% \vspace{-5mm}
\end{table*}

\subsection{Experimental Settings}
To analyze the adaptability of the HGDA under low-resource scenarios, we consider the following experimental settings:
\begin{itemize}
%[leftmargin=*, topsep=0pt, partopsep=0pt,itemsep=0pt,parsep=0pt]
    \item The size of $\mathcal{T}^{'}$: We use $\mathcal{T}^{'} \in \{5, 10, 20, 50\}$ to replicate the data scarcity issue in low-resource scenarios of privately labelled medical corpora.
    \item Sequence encoder adaptation: Following \citet{10.1145/3366423.3380127},
    we consider the hard task of adapting the sequence encoder. This assumes that
    each domain has a domain-specific decoder and only the sentence encoder parameter $\vb*\theta$ is shared across domains and consequently adapted to $\mathcal{T}^{'}$.
\end{itemize}
We implement two variants of HGDA and compare them with the SOTA MetaNER \cite{10.1145/3366423.3380127}. 
\begin{itemize}
%[leftmargin=*, topsep=0pt, partopsep=0pt,itemsep=0pt,parsep=0pt]
    \item \textbf{MetaNER} will act as our major baseline. It is the latest and most related work to HGDA, showing SOTA performance. We followed the parameter settings that the authors detailed in their paper and tried to replicate the MetaNER model based on our understandings. We validated our implementation by comparing its performance to the baseline multi-tasking method used in MetaNER.
    \item \textbf{BioBERT} is used to demonstrate the difficulty that deep PLMs face in low-resource scenarios.
    \item \textbf{HGDA} is one of our set-ups that re-calibrates the gradient updates of $\{\theta,\phi,\omega\}$ using equation~\eqref{eq:TaskScore}.
    \item \textbf{HGDA-NEs} follows the strategy in the task generation discussion, {\it i.e.}, HGDA-NEs only trains with sentences that contains at least one bioNE.
\end{itemize}
As our \text{HGDA} and \text{HGDA-NEs} can either use BiLSTM or BERT as the sequence encoder, we will clearly highlight this information in the presentation of results to avoid any confusions.
Additionally, corpora from the target domain are unseen by the model during the training phase.
For instance, if the ``Disease'' domain is treated as $\mathcal{D}_{\text{target}}$ for adaptation, we only perform learning for $\vb*\Theta^{*}$ using the remaining $\mathcal{D}_{\text{source}} = \{$ ``Drug'', ``Gene'', and ``Species''$\}$. More detailed parameter settings to reproduce this work can be found in the appendix.

\subsection{Results \& Discussions}%Not finished 
~\autoref{tab:heterogeneous} presents the NER performance of MetaNER, BioBERT, HGDA and their variants under the previously defined adaptation settings.
We have the following observations:
\begin{itemize}
%[leftmargin=*, topsep=0pt, partopsep=0pt,itemsep=0pt,parsep=0pt]
    \item MetaNER v.s. HGDA:
    % By including the hardness information in the gradient update, HGDA achieves a significant performance improvement over MetaNER with an average of $4-5\%$ improvement in terms of F1 score.
    % In multiple cases ({\it e.g.}, JNLPBA 5 shots, BC2GM 10 shots, etc), the performance gain of HGDA goes up to 15\% in terms of F1-score. 
    % This comparison demonstrates that using the hardness information to differentiate the importance of each task in the gradient update is beneficial, and contributing largely to the NER performance.
    By simply incorporating the hardness information in the gradient update, HGDA achieves a significant performance improvement over MetaNER with an average of $4-5\%$ improvement in terms of F1 score.
    In multiple cases ({\it e.g.}, JNLPBA 5 shots, BC2GM 10 shots, etc), the performance gain of HGDA goes up to 15\% in terms of F1-score. 
    This result demonstrates that using the hardness to differentiate the importance of each task in the gradient update will contribute to the NER performance.
    \item HGDA v.s. HGDA-NEs:
    Both HGDA and HGDA-NEs work well in our experiments, outperforming the strong baseline by a large margin. HGDA re-weights the gradient update based on the task difficulty and HGDA-NEs trains the learner exclusively only on sentences containing NEs. It is not surprising to see both approaches perform similarly when $\mathcal{T}^{'}$ increases, as HGDA automatically tries to down-weight tasks with sentences containing few/no NEs dynamically.
    \item It is interesting that both HGDA and HGDA-NEs might perform worse than MetaNER on the LINNAEUS corpus. \autoref{tab:Corpora} shows that $87\%$ of LINNAEUS sentences contains no bioNEs. Since both HGDA and HGDA-NEs toss out those sentences implicitly and explicitly during training, this could have attributed to the performance loss.
    \item The BioBERT performance shows the weakness of adapting deep PLMs in the low-resource scenarios for bioNER tasks. 
    Under our HGDA settings, both HGDA* and HGDA-NEs*, which use BERT as the sentence encoder, perform significantly better than the BioBERT baseline. 
    This might suggest that our techniques are architecture invariant.
    Additionally, the significant performance gaps when $\mathcal{T}^{'}$ = \{5,10\} further elevate the necessity of HGDA for deep sentence encoder.
    % only started outperforming our works when there are sufficient instances to train on, {\it e.g.}, 100 sentences. Nonetheless, BioBERT average performance was able to catch up to the other baselines when there were only 50 sentences to train the model. This might imply that under the low-resources setting, our model will perform better; however, when there are sufficient annotated data to train model with, BioBERT should be the preferred system.
\end{itemize}

\noindent Additionally, we also provide the precision and recall results for all of our experiments, these results can be found in the appendix, \autoref{tab:heterogeneous-precision} and \autoref{tab:heterogeneous-recall}.

% \noindent Lastl

%%%%%%%%%%%%%%%%%%%%%%%%%%%%%%%%%%%%%%%%%%%%
% \vspace{-2mm}
\section{Conclusion}
% \vspace{-1mm}
We have proposed simple yet effective methods that effectively leverage the domain hardness information to improve the effectiveness of the learnt model under the low-resource NER settings. Experiments on biomedical corpora have shown that the sequence labelling function derived from our HGDAs have achieved substantial performance improvements compared to current SOTA baselines.
%%%%%%%%%%%%%%%%%%%%%%%%%%%%%%%%%%%%%%%%%%%%

\section*{Limitations}
% \vspace{-2mm}
% EMNLP 2022 requires all submissions to have a section titled ``Limitations'', for discussing the limitations of the paper as a complement to the discussion of strengths in the main text. This section should occur after the conclusion, but before the references. It will not count towards the page limit.  

% The discussion of limitations is mandatory. Papers without a limitation section will be desk-rejected without review.
% ARR-reviewed papers that did not include ``Limitations'' section in their prior submission, should submit a PDF with such a section together with their EMNLP 2022 submission.

% While we are open to different types of limitations, just mentioning that a set of results have been shown for English only probably does not reflect what we expect. 
% Mentioning that the method works mostly for languages with limited morphology, like English, is a much better alternative.
% In addition, limitations such as low scalability to long text, the requirement of large GPU resources, or other things that inspire crucial further investigation are welcome.
HGDA and its variants are trained using English bioNER corpora which have limited morphology. 
We have not applied HGDA to other languages to further verify the performance so this can be a potential area for future works.
To make sure that the batch of tasks are constructed properly, we have to make modifications to the dataloaders. This prevents the GPUs to be fully utilised during training and leads to long training time, {\it e.g.}, taking up to 48 hours to train with 1 NVIDIA RTX3090.
We use multiple RTX3090s to train our models; thus, for GPUs with lower memory, the batch size must be changed which might affect the results.
Since we try to validate the performance of each configuration for 20 times as discussed in the experimental results section, it takes a considerable amount of time to finish the validation of the adaptation performance.
Finally, due to the limitation of available pages, we cannot show detailed information of p-values that suggests the significance of our work.
% \vspace{5mm}
% \vspace{-5mm}
\section*{Ethics Statement}
% \vspace{-2mm}
Our works comply with  \href{https://www.aclweb.org/portal/content/acl-code-ethics}{ACL Ethics Policy}. In this work, we include solely publicly available biomedical corpora that are widely used as benchmarks to measure the bioNER performance and provide proper citations to the authors of these corpora.
% \vspace{-2mm}
% Scientific work published at EMNLP 2022 must comply with the \href{https://www.aclweb.org/portal/content/acl-code-ethics}{ACL Ethics Policy}. We encourage all authors to include an explicit ethics statement on the broader impact of the work, or other ethical considerations after the conclusion but before the references. The ethics statement will not count toward the page limit (8 pages for long, 4 pages for short papers).
% \vspace{5mm}

\bibliography{custom}
\bibliographystyle{acl_natbib}
\clearpage
\appendix
% \vspace{-1mm}
\section{Appendix}
\vspace{-1mm}
\label{sec:appendix}
\begin{algorithm}[t]
    \normalsize
    \caption{HGDA}
    \begin{algorithmic}[1]
        \Require $p(\mathcal{T})$ from source domains
        \Require $\alpha, \beta, \lambda$ hyper-parameters
        \Require m tasks batch size
        \State Initialize $\theta$, $\phi$, $\omega$
        \While{\text{not converge}}
            \For{$i=1,\ldots,\text{m}$}
                \State $\mathcal{T}_i\sim p(\mathcal{T})$% either randomly or sub-section~\ref{subsec:TaskCre}
                \State $\mathcal{T}_i^S, \mathcal{T}_i^Q=\mathcal{T}_i$ s.t. $\mathcal{T}_i^S\cap \mathcal{T}_i^Q=\emptyset$
                \State $\mathcal{L}^{\text{lab}}_i,\mathcal{L}^{\text{cls}}_i =$ algorithm~\ref{alg:Loss}
                \State $\mathcal{L}_i =\mathcal{L}^{\text{lab}}_i+\lambda\mathcal{L}^{\text{cls}}_i$
            \EndFor
            \State $\Gamma_1,\ldots,\Gamma_m= $ equation~\ref{eq:TaskScore} %sub-section~\ref{subsec:TaskHar}%\texttt{Hardness(...)}
            \State $\theta\leftarrow\theta-\alpha\sum_i\gamma_i^{\theta}\nabla_{\theta}\mathcal{L}_i$
            \State $\phi\leftarrow\phi-\alpha\sum_i\gamma_i^{\phi}\nabla_{\phi}\mathcal{L}^{\text{lab}}_i$
            \State $\omega\leftarrow\omega-\alpha\sum_i\gamma_i^{\omega}\nabla_{\omega}\mathcal{L}^{\text{cls}}_i$
        \EndWhile\\
        \Return $\Theta =\left(\theta,\phi\right)$
    \end{algorithmic}
    \label{alg:MetaBioNER}
    \vspace{-1mm}
\end{algorithm}

\textbf{Detailed experimental setups}
HGDA and HGDA-NEs can use either BiLSTM or BERT as the sentence encoder. 
When BiLSTM is used as the sentence encoder, we use CRF as the tag decoder and a fully connected layer as the domain classifier.
The size of the token embeddings from BioWordVec is 200. 
Aside from using this token embeddings to feed to the sentence encoder, we also have one LSTM and one CNN network to learn the character embeddings with the output of 50. 
Combining the token embeddings and the character embeddings gives the input of 300 to the BiLSTM sentence encoder. 
The output of this BiLSTM is 256 (128*2), this output is then fed to two separate heads in the network. 
One of them is the CRF tag-decoder which generate the BIO tagging sequence with with B-Begin, I-Inside, and O-Outside of NEs. 
The other is a fully connected layer that predicts which domain the sentences in $\mathcal{T}_i$ belong to. During training, we set the default learning rate of 1e-2 for both  $\alpha$ and $\beta$. 
As this is the preferred learning rate for the BiLSTM with a batch size of 32, these learning rates are subjected to changes depending on the training batch-size, {\it i.e.}, $K$ and $\mathcal{T}^{'}$ as previously discussed. 
For each $K$, we calculate $\alpha$ and $\beta$ using
\begin{equation}
    \small
    \label{eq:alpha}
    \alpha=\beta=\text{Default Learning Rate}*\sqrt{\frac{K}{32}}
    \vspace{-1mm}
\end{equation}
When BERT \cite{devlin-etal-2019-bert,wolf-etal-2020-transformers} acts as the sequence encoder, we set the max sequence length for padding and truncating to be 256 as biomedical texts tends to be longer than the general texts.
We use cased vocabulary for a slightly better performance and set the  dimensionality of the encoder layers and the pooler layer to 768.
For the tokenization, BERT uses WordPiece tokenization \cite{wu2016google} to deal with the out-of-vocabulary (OOV) issue which is common for biomedical texts. 
The default learning rates $\alpha$ and $\beta$ for $K$ of 32 are set to 1e-5 and are subjected to changes as shown in Eq.~\ref{eq:alpha}. The output from the BERT sequence encoder will then be fed into two separate fully connected layers. One of them is to predict the tagging sequence. The other is to predict which domain for the sentences in $\mathcal{T}_i$.

\vspace{9pt}
\vspace{45535sp}
All models are trained using the SGD optimizer \cite{kiefer1952stochastic} with a linear learning rate scheduler \cite{wolf-etal-2020-transformers}. We set the gradient clip at 5; momentum at 0.9; weight decay at 1e-6; and dropout rate at 0.2 for all our training \cite{10.1145/3366423.3380127}. After cross-validating for different values of $\lambda$ , we use $\lambda=1$ for all HGDA variants to control the trade-off between the sequence labelling loss and the domain classifying loss. Additionally, since HGDA involves the bilevel optimization framework, we have to approximate for the gradients acquired from $\mathcal{T}_{i}^{Q}$ using first order gradient approximations \cite{DBLP:journals/corr/abs-1803-02999} and implicit gradients \cite{rajeswaran2019meta} to avoid the computation for the Jacobian matrix.
%We plan to release 
Please contact the corresponding author for the codes to re-implement this work.
\begin{algorithm}[t]
    \normalsize
    \caption{Bilevel optimization for $\mathcal{T}_i$}
    \begin{algorithmic}[1]
        \Require $\mathcal{T}_i=\left(\mathcal{T}_i^S, \mathcal{T}_i^Q\right)$ s.t. $\mathcal{T}_i^S\cap \mathcal{T}_i^Q=\emptyset$
        \Require $\theta, \phi, \omega$ current iteration parameters
        \Require $\beta,\lambda$ hyper-parameters
        \State Initialize $\theta_i$, $\phi_i$, $\omega_i$ with $\theta, \phi, \omega$
        \For {$ i = 1,\ldots,\text{adaptation steps}$} %K$}
            \State $\mathcal{L}^{\text{lab}}_i = \mathcal{L}\left(h\left(\theta_i,\phi_i\right),\mathcal{T}_i^S\right)$
            \State $\mathcal{L}^{\text{cls}}_i = \mathcal{L}\left(f\left(\theta_i,\omega_i\right),\mathcal{T}_i^S\right)$
            \State $\mathcal{L}_i =\mathcal{L}^{\text{lab}}_i+\lambda\mathcal{L}^{\text{cls}}_i$
            \State $\theta_i\leftarrow\theta_i-\beta\nabla_{\theta_i}\mathcal{L}_i$
            \State $\phi_i\leftarrow\phi_i-\beta\nabla_{\phi_i}\mathcal{L}^{\text{lab}}_i$
            \State $\omega_i\leftarrow\omega_i-\beta\nabla_{\omega_i}\mathcal{L}^{\text{cls}}_i$
        \EndFor
        \State $\mathcal{L}^{\text{lab}}_i = \mathcal{L}\left(h\left(\theta_i,\phi_i\right),\mathcal{T}_i^Q\right)$
        \State $\mathcal{L}^{\text{cls}}_i = \mathcal{L}\left(f\left(\theta_i,\omega_i\right),\mathcal{T}_i^Q\right)$
        \\\Return $\mathcal{L}^{\text{lab}}_i,\mathcal{L}^{\text{cls}}_i$
    \end{algorithmic}
    \label{alg:Loss}
    \vspace{3mm}
\end{algorithm}

\begin{table*}[!th]
\centering
\small
\textbf{Precision Performance}
\begin{adjustbox}{max width=\textwidth}
\begin{tabular}{|c|l|cc|cc|cc|cc|}
    \hline
    \multicolumn{1}{|c|}{$\mathcal{T}^{'}$}
        & \multicolumn{1}{c|}{}
        & \multicolumn{2}{c|}{Disease} 
        & \multicolumn{2}{c|}{Drug} 
        & \multicolumn{2}{c|}{Gene} 
        & \multicolumn{2}{c|}{Species}
        % & \multicolumn{1}{c|}{Overall}
    \\ \cline{3-10}
    
    \multicolumn{1}{|c|}{Size}
        & \multicolumn{1}{c|}{}
        & NCBI & BC5CDR 
        & BC5CDR & BC4CHEMD 
        & JNLPBA  & BC2GM
        & LINNAEUS  & S800 
        
    \\ \hline 
    
    \parbox[t]{3mm}{\multirow{6}{*}{\rotatebox[origin=c]{90}{\textbf{5}}}}
    &\textbf{MetaNER}
        &{0.3645}&{0.4010}&{0.7740}&{0.4048}&{0.1717}&{0.2050}&{0.6278}&{0.2589}\\

    &\textbf{BioBERT}
        &{0.0324}&{0.0305}&{0.1099}&{0.0283}&{0.0573}&{0.0489}&{0.0055}&{0.0054}\\
        
    &\textbf{HGDA}
        &{0.3937}&{0.5039}&{0.7355}&{0.4085}&{0.3175}&{0.3431}&{0.4312}&{0.3538}\\
    
    &\textbf{HGDA-NEs}
        &{0.4540}&{0.5044}&{0.7023}&{0.3203}&{0.3507}&{0.3887}&{0.5784}&{0.4551}\\

    &\textbf{HGDA*}
        &{0.2219}&{0.1419}&{0.5400}&{0.1502}&{0.2928}&{0.2616}&{0.0860}&{0.1899}\\
    
    &\textbf{HGDA-NEs*}
        &{0.3059}&{0.2203}&{0.5858}&{0.1761}&{0.2546}&{0.2571}&{0.1954}&{0.2401}\\
    
    \hline \hline
    
    \parbox[t]{3mm}{\multirow{6}{*}{\rotatebox[origin=c]{90}{\textbf{10}}}}
    &\textbf{MetaNER}
        &0.3176&0.3721&0.7079&0.3398&0.2864&{0.3175}&{0.6281}&{0.3812}\\
    
    &\textbf{BioBERT}
        &{0.1038}&{0.0426}&{0.3964}&{0.0916}&{0.1608}&{0.2480}&{0.0444}&{0.0221}\\
        
    &\textbf{HGDA}
        &{0.4311}&{0.4656}&{0.7227}&{0.4457}&{0.4300}&{0.4468}&{0.4991}&{0.3714}\\

    &\textbf{HGDA-NEs}
        &{0.5715}&{0.5651}&{0.7079}&{0.4226}&{0.3762}&{0.3938}&{0.4354}&{0.3206}\\

    &\textbf{HGDA*}
        &{0.3318}&{0.3581}&{0.6324}&{0.2844}&{0.3370}&{0.3117}&{0.1264}&{0.2909}\\
    
    &\textbf{HGDA-NEs*}
        &{0.3847}&{0.3172}&{0.6815}&{0.3666}&{0.3113}&{0.2824}&{0.1452}&{0.2844}\\
    
    \hline \hline
        
    \parbox[t]{3mm}{\multirow{6}{*}{\rotatebox[origin=c]{90}{\textbf{20}}}}
    &\textbf{MetaNER}
        &{0.4699}&{0.4652}&{0.7523}&{0.4384}&{0.3516}&{0.3625}&{0.5462}&{0.2611}\\
    
    &\textbf{BioBERT}
        &{0.3907}&{0.3402}&{0.7389}&{0.2518}&{0.3048}&{0.4067}&{0.1954}&{0.2362}\\
    
    &\textbf{HGDA}
        &{0.5968}&{0.5926}&{0.7846}&{0.4771}&{0.4836}&{0.4910}&{0.5205}&{0.4443}\\

    &\textbf{HGDA-NEs}
        &{0.5698}&{0.5263}&{0.7650}&{0.4680}&{0.4745}&{0.4805}&{0.7638}&{0.5419}\\
    
    &\textbf{HGDA*}
        &{0.3634}&{0.3876}&{0.6338}&{0.3660}&{0.3918}&{0.3467}&{0.1243}&{0.2787}\\
    
    &\textbf{HGDA-NEs*}
        &{0.4354}&{0.4587}&{0.7027}&{0.4173}&{0.3722}&{0.3428}&{0.2455}&{0.3205}\\
        
    \hline \hline
        
    \parbox[t]{3mm}{\multirow{6}{*}{\rotatebox[origin=c]{90}{\textbf{50}}}}
        
    &\textbf{MetaNER}
        &{0.6154}&{0.6150}&{0.7500}&{0.4824}&{0.4938}&{0.5030}&{0.6400}&{0.3455}\\
    
    &\textbf{BioBERT}
        &{0.5661}&{0.5455}&{0.7679}&{0.4308}&{0.3963}&{0.5314}&{0.6914}&{0.4020}\\
    
    &\textbf{HGDA}
        &{0.6659}&{0.5916}&{0.8006}&{0.5734}&{0.5035}&{0.5141}&{0.7683}&{0.5911}\\
    
    &\textbf{HGDA-NEs}
        &{0.6320}&{0.5692}&{0.7722}&{0.5585}&{0.5587}&{0.5901}&{0.7331}&{0.5578}\\
    
    &\textbf{HGDA*}
        &{0.5060}&{0.5258}&{0.7083}&{0.3758}&{0.4202}&{0.3820}&{0.3580}&{0.3699}\\

    &\textbf{HGDA-NEs*}
        &{0.5536}&{0.5720}&{0.7566}&{0.5497}&{0.4971}&{0.4697}&{0.3243}&{0.3663}\\
    \hline
    \end{tabular}
\end{adjustbox}
% \vspace{-2mm}
\caption{
Average precision-performance of the sequence encoder adaptation for bioNER tasks with the best performance boldfaced. All results have the same settings with those from \autoref{tab:heterogeneous}.
}
\label{tab:heterogeneous-precision}
% \vspace{-5mm}
\end{table*}

\begin{table*}[!th]
\centering
\small
\textbf{Recall Performance}
\begin{adjustbox}{max width=\textwidth}
\begin{tabular}{|c|l|cc|cc|cc|cc|}
    \hline
    \multicolumn{1}{|c|}{$\mathcal{T}^{'}$}
        & \multicolumn{1}{c|}{}
        & \multicolumn{2}{c|}{Disease} 
        & \multicolumn{2}{c|}{Drug} 
        & \multicolumn{2}{c|}{Gene} 
        & \multicolumn{2}{c|}{Species}
        % & \multicolumn{1}{c|}{Overall}
    \\ \cline{3-10}
    
    \multicolumn{1}{|c|}{Size}
        & \multicolumn{1}{c|}{}
        & NCBI & BC5CDR 
        & BC5CDR & BC4CHEMD 
        & JNLPBA  & BC2GM
        & LINNAEUS  & S800 
        % &    
    \\ \hline 
    
    \parbox[t]{3mm}{\multirow{6}{*}{\rotatebox[origin=c]{90}{\textbf{5}}}}
    &\textbf{MetaNER}
        &{0.2493}&{0.1689}&{0.4818}&{0.1725}&{0.3099}&{0.3196}&{0.0700}&{0.1110}\\

    &\textbf{BioBERT}
        &{0.1029}&{0.1294}&{0.0828}&{0.0369}&{0.1402}&{0.0700}&{0.0932}&{0.0482}\\
        
    &\textbf{HGDA}
        &{0.2780}&{0.2084}&{0.5446}&{0.2027}&{0.4580}&{0.3454}&{0.1212}&{0.2498}\\
    
    &\textbf{HGDA-NEs}
        &{0.2247}&{0.1934}&{0.4852}&{0.2471}&{0.4280}&{0.3467}&{0.0947}&{0.1981}\\

    &\textbf{HGDA*}
        &{0.2617}&{0.0485}&{0.4759}&{0.1251}&{0.5114}&{0.4327}&{0.0719}&{0.4199}\\
    
    &\textbf{HGDA-NEs*}
        &{0.3342}&{0.1048}&{0.6442}&{0.3728}&{0.4824}&{0.4504}&{0.0498}&{0.3829}\\
    
    \hline \hline
    
    \parbox[t]{3mm}{\multirow{6}{*}{\rotatebox[origin=c]{90}{\textbf{10}}}}
    &\textbf{MetaNER}
        &{0.3555}&{0.3913}&{0.6381}&{0.3574}&{0.4149}&{0.3447}&{0.2052}&{0.2843}\\
    
    &\textbf{BioBERT}
        &{0.1275}&{0.0970}&{0.2724}&{0.0997}&{0.2612}&{0.1973}&{0.1266}&{0.0728}\\
        
    &\textbf{HGDA}
        &{0.3887}&{0.3904}&{0.6486}&{0.4135}&{0.5497}&{0.4615}&{0.2134}&{0.3802}\\

    &\textbf{HGDA-NEs}
        &{0.3627}&{0.3481}&{0.6261}&{0.3996}&{0.5313}&{0.4334}&{0.1821}&{0.3341}\\

    &\textbf{HGDA*}
        &{0.4631}&{0.4704}&{0.7061}&{0.4870}&{0.5799}&{0.5500}&{0.1883}&{0.5045}\\
    
    &\textbf{HGDA-NEs*}
        &{0.4503}&{0.3216}&{0.7438}&{0.4740}&{0.5507}&{0.5459}&{0.1255}&{0.4733}\\
    
    \hline \hline
        
    \parbox[t]{3mm}{\multirow{6}{*}{\rotatebox[origin=c]{90}{\textbf{20}}}}
    &\textbf{MetaNER}
        &{0.4559}&{0.4859}&{0.7116}&{0.4460}&{0.5133}&{0.4325}&{0.4574}&{0.3574}\\
    
    &\textbf{BioBERT}
        &{0.3126}&{0.2429}&{0.5582}&{0.2351}&{0.4923}&{0.4529}&{0.1290}&{0.2422}\\
    
    &\textbf{HGDA}
        &{0.5362}&{0.5235}&{0.7167}&{0.5194}&{0.6315}&{0.5391}&{0.2922}&{0.4449}\\

    &\textbf{HGDA-NEs}
        &{0.5418}&{0.5019}&{0.7009}&{0.4768}&{0.6135}&{0.5573}&{0.3598}&{0.5076}\\
    
    &\textbf{HGDA*}
        &{0.5406}&{0.6074}&{0.7846}&{0.5911}&{0.6426}&{0.5866}&{0.2584}&{0.5503}\\
    
    &\textbf{HGDA-NEs*}
        &{0.5395}&{0.5945}&{0.7717}&{0.6138}&{0.6390}&{0.6015}&{0.3563}&{0.5545}\\
        
    \hline \hline
        
    \parbox[t]{3mm}{\multirow{6}{*}{\rotatebox[origin=c]{90}{\textbf{50}}}}
        
    &\textbf{MetaNER}
        &{0.5382}&{0.6079}&{0.7475}&{0.5392}&{0.5825}&{0.5109}&{0.5918}&{0.3819}\\
    
    &\textbf{BioBERT}
        &{0.6406}&{0.6108}&{0.7384}&{0.5691}&{0.6283}&{0.6608}&{0.5122}&{0.5379}\\
    
    &\textbf{HGDA}
        &{0.5896}&{0.5982}&{0.7499}&{0.5732}&{0.6486}&{0.5792}&{0.5480}&{0.5830}\\
    
    &\textbf{HGDA-NEs}
        &{0.6115}&{0.6027}&{0.7541}&{0.6024}&{0.6845}&{0.6149}&{0.5670}&{0.5361}\\
    
    &\textbf{HGDA*}
        &{0.6333}&{0.6690}&{0.8172}&{0.5644}&{0.7025}&{0.6429}&{0.5883}&{0.6043}\\
    
    &\textbf{HGDA-NEs*}
        &{0.6475}&{0.6764}&{0.8554}&{0.7023}&{0.7095}&{0.6563}&{0.6823}&{0.6508}\\
    \hline
    \end{tabular}
\end{adjustbox}
% \vspace{-2mm}
\caption{
Average recall-performance of the sequence encoder adaptation for bioNER tasks with the best performance boldfaced. All results have the same settings with those from \autoref{tab:heterogeneous}.
}
\label{tab:heterogeneous-recall}
% \vspace{-5mm}
\end{table*}

\end{document}